\documentclass{article}
\PassOptionsToPackage{numbers, compress}{natbib}

\usepackage[final]{nips_2016}
\usepackage[utf8]{inputenc} 
\usepackage[T1]{fontenc}    
\usepackage{hyperref}       
\usepackage{geometry}                		
\usepackage{amssymb}

\usepackage{url}            
\usepackage{booktabs}       
\usepackage{amsmath,amssymb,mathtools,bm,etoolbox}
\usepackage{amsfonts}       
\usepackage{nicefrac}       
\usepackage{microtype}
\usepackage{graphicx}      
\graphicspath{{Images/}}
\usepackage[autostyle]{csquotes}
\usepackage{multirow}

\title{Learning what to look in chest X-rays with a recurrent visual attention model}

\author{
  Petros-Pavlos Ypsilantis \\
  Department of Biomedical Engineering\\
  King's College London\\
  London, SE1 7EH \\
  \texttt{petros-pavlos.ypsilantis@kcl.ac.uk} \\
  \And
  Giovanni Montana \\
  Department of Biomedical Engineering\\
  King's College London \\
  London, SE1 7EH \\
  \texttt{giovanni.montana@kcl.ac.uk} 
}

\begin{document}

\maketitle

\begin{abstract}

X-rays are commonly performed imaging tests that use small amounts of radiation to produce pictures of the organs, tissues, and bones of the body. X-rays of the chest are used to detect abnormalities or diseases of the airways, blood vessels, bones, heart, and lungs. In this work we present a stochastic attention-based model that is capable of learning what regions within a chest X-ray scan should be visually explored in order to conclude that the scan contains a specific radiological abnormality. The proposed model is a recurrent neural network (RNN) that learns to sequentially sample the entire X-ray and focus only on informative areas that are likely to contain the relevant information. We report on experiments carried out with more than $100,000$ X-rays containing enlarged hearts or medical devices. The model has been trained using reinforcement learning methods to learn task-specific policies.  
   
\end{abstract}

\section{Introduction}

Chest X-rays (CXR) are the most commonly used diagnosis exams for chest-related diseases. They use a very small dose of ionizing radiation to produce pictures of the inside of the chest. CXR scans help radiologists to diagnose or monitor treatment for conditions such as pneumonia, heart failure, emphysema, lung cancer, positioning of medical devices, as well as fluid and air collection around the lungs. An expert radiologist is typically able to detect radiological abnormalities by looking in the 'right places'"' and making quick comparisons to normal standards. For example, for the detection of an enlarged heart, or cardiomegaly, the size of the heart is assessed in relation to the total thoracic width. Given that chest X-rays are routinely used to detect several abnormalities or diseases, a careful interpretation of a scan requires expertise and time resources that are not always available, especially since large numbers of CXR exams need to be reported daily. This leads to diagnostic errors that, for some pathologies, have been estimated to be in the range of $23\%$ \cite{Tudor97}.

Our ultimate objective is to develop a fully-automated system that learns to identify radiological abnormalities using only large volumes of labelled historical exams. We are motivated by recent work on attention-based models which have been used for digit classification \cite{Mnih14}, sequential prediction of street view house numbers \cite{Ba15} and a fine-grained categorization task \cite{Sermanet15}. However, we are not aware of applications of such attention models to the challenging task of chest X-ray interpretation. Here we report on the initial performance of a recurrent attention model (RAM), similar to the model originally presented in \cite{Mnih14}, and trained end-to-end on a very large number of historical X-rays exams. 

\section{Dataset}

For this study we collected and prepared a dataset consisting of $745,480$ X-ray plain films of the chest along with their corresponding radiological reports. All the historical exams were extracted from the historical archives of Guy's and St Thomas' Hospital in London (UK), and covered more than a decade, from $2005$ to $2016$. Each scan was labelled according to the clinical findings that were originally reported by the consultant radiologist and recorded in an electronic clinical report. The labelling task was automated using a natural language processing (NLP) system that implements a combination of machine learning and rule-based algorithms for clinical entity recognition, negation detection and entity classification. An early version of the system used a bidirectional long-short term memory (LSTM) model for modelling the radiological language and detecting clinical findings and their negations \cite{Cornegruta16}.

For the purpose of this study, we only used scans labelled as {\it normal} (i.e. those with no reported abnormalities), and those reported as having an enlarged heart (i.e. a large cardiac silhouette) and a medical device (e.g. a pacemaker). The number of scans within these three categories was $57,595$, $31,528$ and $22,804$, respectively. We were interested in the detection of enlarged hearts and medical devices, and a separate model was trained for each task and tested on $6,000$ and $4,000$ randomly selected exams, respectively. All the remaining images were used for both training and validation. In all our experiments we scaled the size of the images down to $256 \times 256$ pixels.
   
\section{Recurrent attention model (RAM)}

The RAM model implemented here is similar to the one originally proposed in \cite{Mnih14}. Mimicking the human visual attention mechanism, the this model learns to focus and process only a certain region of an image that is relevant to the classification task. In this section we provide a brief overview of the model and describe how our implementation differs from the original architecture. We refer the reader to \cite{Mnih14} for further details on the training algorithm.

\begin{figure}[t]
\begin{center}
\includegraphics[width=4.6in]{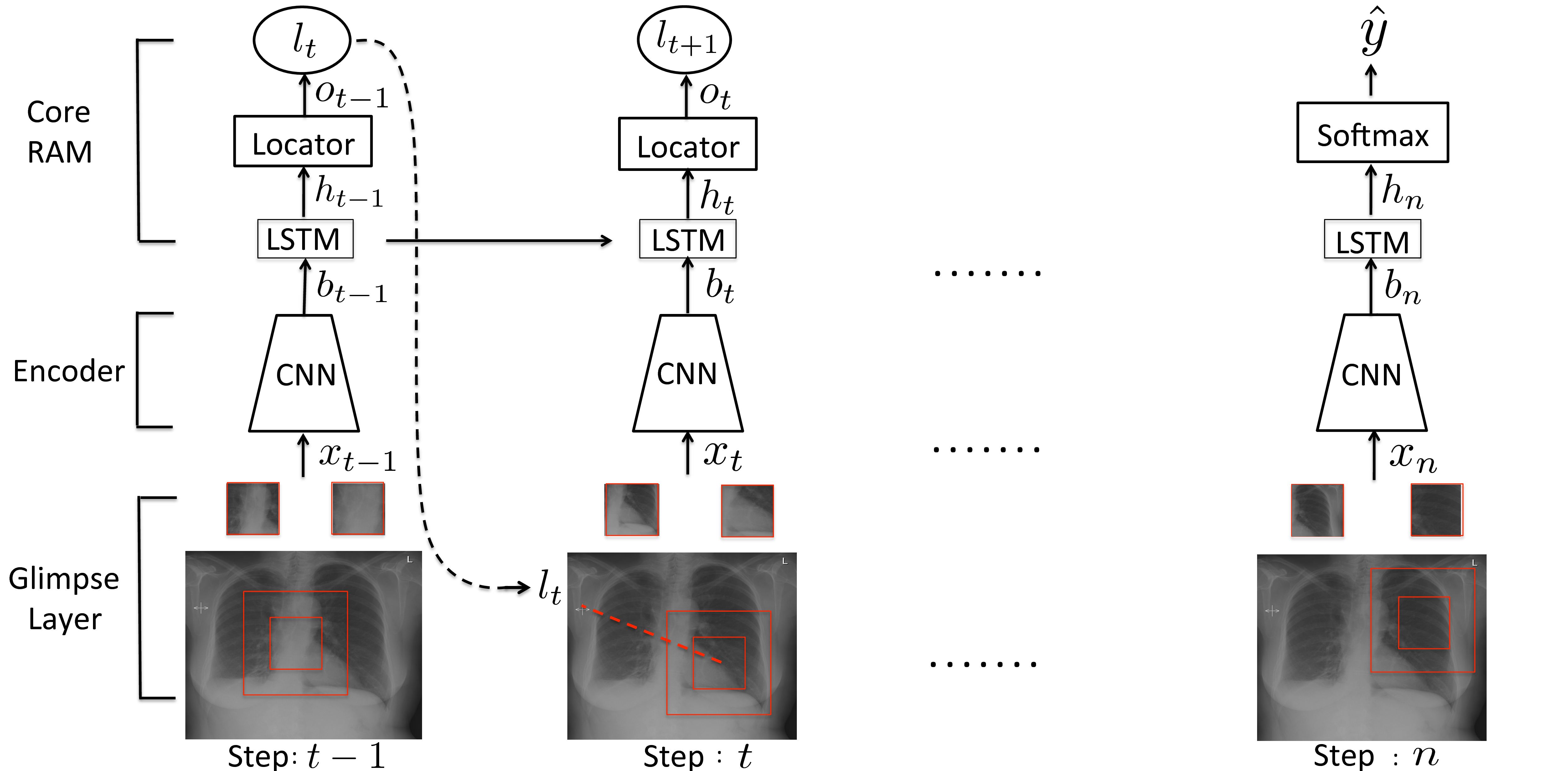}
\end{center}
\caption{RAM. At each time step $t$ the Core RAM samples a location $l_{t+1}$ of where to attend next. The location $l_{t+1}$ is used to extract the glimpse (red frames of different size). The  image patches are down-scaled and passed through the encoder. The representation of the encoder and the previous hidden state $h_{t-1}$ of the Core RAM are passed as inputs to the LSTM of the step $t$. The locator receives as input the hidden state $h_{t}$ of the current LSTM and then it samples the location coordinates for the glimpse in the next step $t+1$. This process continuous recursively until step $n$ where the output of the LSTM $h_{n}$ is used to classify the input image.}
\label{fig:ram}
\end{figure}

{\bf Glimpse Layer:} At each time $t$, the model does not have full access to the input image but instead receives a partial observation, or \enquote{glimpse}, denoted by ${\bf x}_{t}$. The glimpse consists of two image patches of different size centred at the same location ${\bf l}_{t}$, each one capturing a different context around ${\bf l}_{t}$. Both patches are matched in size and passed as input to an encoder, as illustrated in Figure \ref{fig:ram}. 

{\bf Encoder:} The encoder implemented here differs from the one used in \cite{Mnih14}. In our application we have a complex visual environment featuring high variability in both luminance and object complexity. This is due to the large variability in patient's' anatomy as well as image acquisition as the X-ray scans were aquired using more than $20$ different X-ray devices. The goal of the encoder is to compress the information of the glimpse by extracting a robust representation. To achieve this, each image of the glimpse is passed through a stack of two convolutional autoencoders with max-pooling \citep{Masci11}. Each convolutional autoencoder in the stack is pre-trained separately from the RAM model. During training, at each time $t$ the glimpse representation is concatenated with the location representation and passed as input to a fully connected (FC) layer. The output of the FC layer is denoted as ${\bf b}_{t}$ and is passed as input to the core RAM model, as seen in Figure \ref{fig:ram}.

{\bf Core RAM:} In each time step $t$, the output vector ${\bf b}_{t}$ and the previous hidden representation ${\bf h}_{t-1}$ are passed as input to the LSTM layer. The locator receives the hidden representation ${\bf h}_{t}$ from the LSTM unit and passes on to a FC layer, resulting in a vector ${\bf o}_{t} \in R^{2}$ (see Figure \ref{fig:ram}). The locator then decides the position of the next glimpse by sampling ${\bf l}_{t+1} \sim N({\bf o}_{t}, \Sigma)$, i.e. from a normal distribution with mean ${\bf o}_{t}$ and diagonal covariance matrix $\Sigma$. The location ${\bf l}_{t+1}$ represents the $\texttt{x}$-$\texttt{y}$ coordinates of the glimpse at time step $t+1$. At the very first step, we initiate the algorithm at the center of the image, and always use a fixed variance $\sigma^{2}$.  

\section{Results}

Table 1 summaries the classification performance of the RAM model alongside with the performance of state-of-the-art convolutional neural networks trained and tested on the same dataset. RAM, using $5$ million parameters, reaches $90.6\%$ and $91.0\%$ accuracy for the detection of medical devices and enlarged hearts, respectively. For the same tasks, Inception-v3 \cite{Szegedy15} achieves the highest accuracy with $91.3\%$ and $91.4\%$, but uses $4$ times more parameters compared to the RAM model.

\begin{table*}[!htbp]
\caption{Accuracy ($\%$) on the classification between images with normal radiological appearance and images with enlarged heart and medical devices.}
\label{tbl:results}
\begin{center}
\begin{tabular}{|c|c|c|c|c|}
\hline 
\hline 
 \textbf{Model} & \textbf{Heart Enlarged} & \textbf{Medical Devices} & \textbf{Number of Parameters}\\
\hline 
VGG \cite{Simonyan15} & $89.1$ & $86.2$ & $\sim 68$ million \\
\hline
ResNet-18 \cite{He15} & $89.5$ & $ 88.4$ & $\sim 11$ million \\
\hline
Inception-v3 \cite{Szegedy15} & ${\bf 91.4}$ & ${\bf 91.3}$ & $\sim 21$ million \\

\hline

AlexNet \cite{Krizhevsky12} & $89.7$ & $87.4$ & $\sim 70$ million \\

\hline

RAM &  $91.0$& $90.6$ &  $\sim 5$ million \\

\hline

baseline ResNet-18 & $90.0$ & $87.2$ & $\sim 5$ million \\

\hline

baseline VGG & $88.9$ & $86.3$ & $\sim 5$ million  \\

\hline

baseline AlexNet & $90.0$ & $86.3$ & $\sim 5$ million \\

\hline
\end{tabular}
\end{center}
\end{table*}

In Figure \ref{fig:attention_object} we illustrate the performance on the validation set, and highlight the locations attended by the model when trying to detect medical devices. Here it can be noted how initially the model explores randomly selected portions of the image, and its classification performance remains low. After a specific number of epochs, the model discovers that the most informative parts of a chest X-ray are those containing the lungs and spine, and selectively prefers those regions in subsequent paths. This is a reasonable policy since most of the medical devices to be found in chest X-rays, such as pacemakers and tubes, are located in those areas.

\begin{figure}[!htbp]
\begin{center}
\includegraphics[width=4.8in]{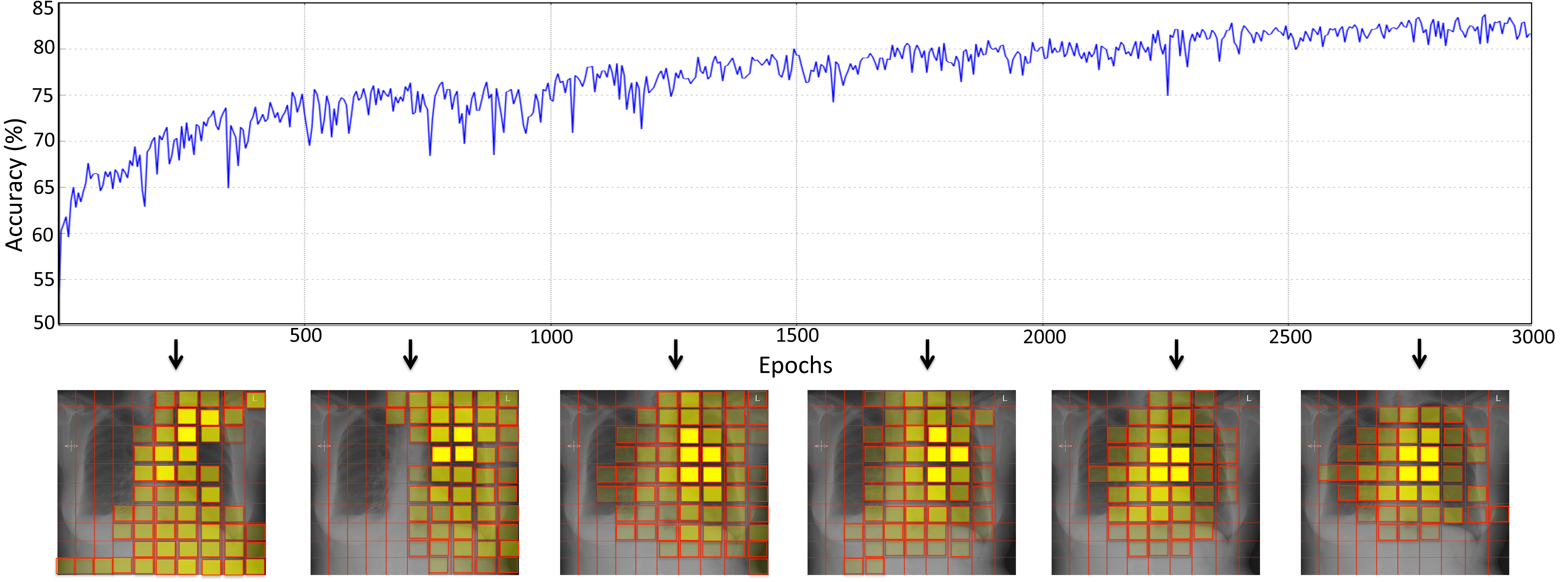}
\end{center}
\caption{Top: Accuracy ($\%$) on the validation set during the training of the model. Bottom: Image locations that the model attends during the validation. The grids corresponds to image areas of size $25 \times 25$ pixels. We split the total number of epochs into chunks of $500$ epochs. For each chunk the model is validated $100$ times and the locations from each validation are summarized in the corresponding image. Image regions with low transparency correspond to locations that the model visits with high frequency.}
\label{fig:attention_object}
\end{figure}

Figure \ref{fig:attention_object} (A) shows the locations mostly attended by the RAM model when looking for medical devices. From this figure it is obvious that the learnt policy explores only the relevant areas where these devices can generally be found. Two examples of paths followed by the algorithm after learning the policy are illustrated in Figures \ref{fig:attention_object} (B), (C). In these examples, starting from the center of the image, the algorithms moves closer to a region that is likely to contain a pacemaker, which is then correctly identified. The circle and triangle points (in red) indicate the coordinates of the first and last glimpse in the learnt policy, respectively.

\begin{figure}[!htbp]
\begin{center}
\includegraphics[width=4.5in]{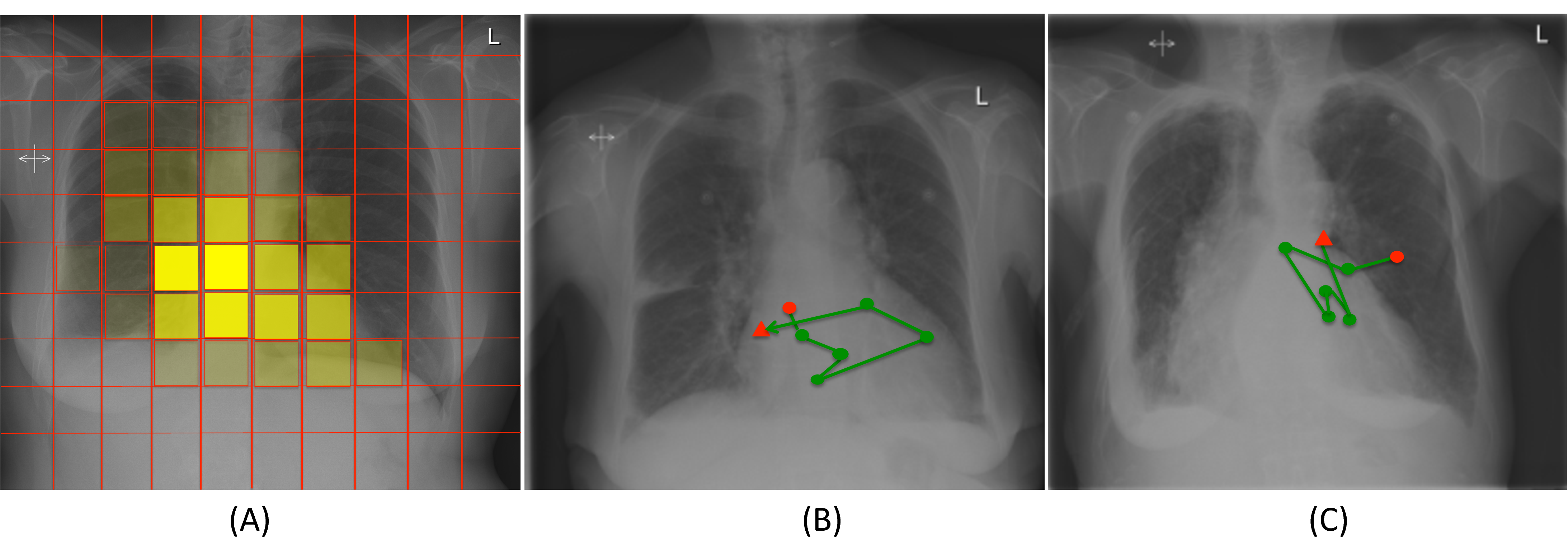}
\end{center}
\caption{(A) Image locations attended by the RAM model for the detection of enlarged hearts. (B) and (C) are two different samples of the learnt policy on test images.}
\label{fig:attention_heart_enlarged}
\end{figure}

Analogously, Figure~\ref{fig:attention_heart_enlarged} (A) highlights frequently explored locations when trying to discriminate between normal and enlarged hearts. Here it can be observed how the models learns to focus on the cardiac area. Two samples of the learned policy are illustrated in Figure~\ref{fig:attention_heart_enlarged} (B), (C). The trajectories followed here demonstrates how the policy has learned that exploring the extremities of the heart is required in order to conclude whether the heart is enlarged or not. 

\begin{figure}[!htbp]
\begin{center}
\includegraphics[width=4.5in]{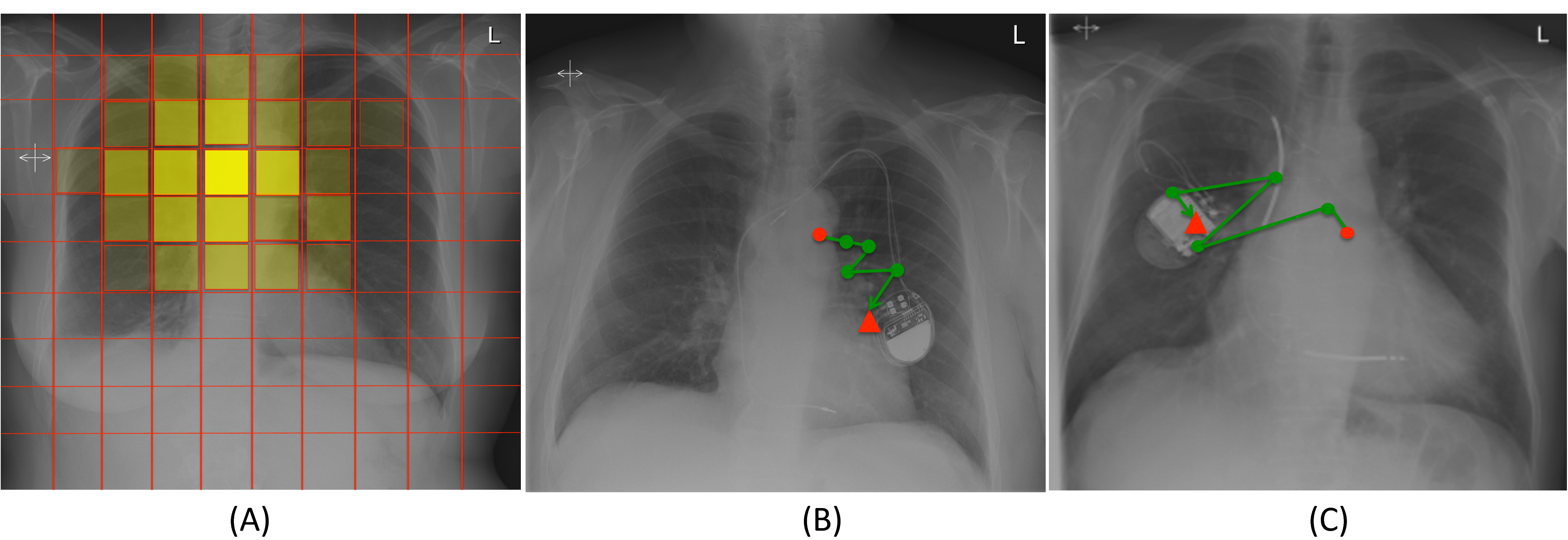}
\end{center}
\caption{(A) Image locations attended by the RAM model for the detection of medical devices. (B) and (C) are two different samples of the learnt policy on test images.}
\label{fig:attention_object}
\end{figure}

\section{Conclusion and Perspectives}

In this work we have investigated whether a visual attention mechanism, the RAM model, is capable of learning how to interpret chest X-ray scans. Our experiments show that the model not only has the potential to achieve classifcation performance comparable to state-of-the-art convolutional architectures using far fewer parameters, but also learns to identify specific portions of the images that are likely to contain the anatomical information required to reach correct conclusions. The relevant areas are explored according to policies that seem appropriate for each task. Current work is being directed towards enabling the model to learn each policy as quickly and precisely as possible using full-scale images and for a much larger number of clinically important radiological classes.

\bibliography{shot_notes}
\bibliographystyle{splncsnat}

\end{document}